\documentclass[journal]{IEEEtran}
\usepackage{filecontents}

\usepackage{graphicx}
\usepackage{cite}
\usepackage{amsmath}
\usepackage{amssymb}
\usepackage{lipsum}
\usepackage{subfig}
\usepackage{multirow}

\ifCLASSINFOpdf
\else
\fi

\begin{document}

\title{Voxel-wise Adversarial Semi-supervised Learning for Medical Image Segmentation}
\author{Chae Eun Lee$^{\dagger}$, Hyelim Park$^{\dagger}$, Yeong-Gil Shin, and Minyoung Chung$^{\ast}$
\thanks{\textit{$^{\dagger}$ Both authors contributed equally to the work.}}%
\thanks{\textit{$^{\ast}$ Corresponding author.}}%
\thanks{This work was supported by Institute of Information \& communications Technology Planning \& Evaluation (IITP) grant funded by the Korea government(MSIT) (No.2021-0-00511, Robust AI and Distributed Attack Detection for Edge AI Security).}
\thanks{C. Lee is with the Samsung Electronics, South Korea (e-mail: chaenn.lee@samsung.com).}
\thanks{H. Park and Y.-G. Shin are with the Department of Computer Science and Engineering, Seoul National University, South Korea.}%
\thanks{*M. Chung is with the School of Software, Soongsil University, South Korea (e-mail: chungmy@ssu.ac.kr).}
}

 
\maketitle

\begin{abstract}
Semi-supervised learning for medical image segmentation is an important area of research for alleviating the huge cost associated with the construction of reliable large-scale annotations in the medical domain. Recent semi-supervised approaches have demonstrated promising results by employing consistency regularization, pseudo-labeling techniques, and adversarial learning. These methods primarily attempt to learn the distribution of labeled and unlabeled data by enforcing consistency in the predictions or embedding context. However, previous approaches have focused only on local discrepancy minimization or context relations across single classes. In this paper, we introduce a novel adversarial learning-based semi-supervised segmentation method that effectively embeds both local and global features from multiple hidden layers and learns context relations between multiple classes. Our voxel-wise adversarial learning method utilizes a voxel-wise feature discriminator, which considers multilayer voxel-wise features (involving both local and global features) as an input by embedding class-specific voxel-wise feature distribution. Furthermore, we improve our previous representation learning method by overcoming information loss and learning stability problems, which enables rich representations of labeled data. Our method outperforms current best-performing state-of-the-art semi-supervised learning approaches on the image segmentation of the left atrium (single class) and multiorgan datasets (multiclass). Moreover, our visual interpretation of the feature space demonstrates that our proposed method enables a well-distributed and separated feature space from both labeled and unlabeled data, which improves the overall prediction results.
\end{abstract}


\IEEEpeerreviewmaketitle

\section{Introduction}

\IEEEPARstart{M}{edical} image segmentation is an essential task in several clinical approaches, such as computer-aided diagnosis, radiation therapy, and virtual surgeries \cite{788580,robotics_surgery,cad}. Automated segmentation of organs (e.g., left atrium (LA), heart, or liver) is of significant importance in optimizing clinical workflow, such as the planning of surgeries and treatments. Convolutional neural networks (CNNs), which have demonstrated significant abilities in learning visual features in computer vision tasks, have been successfully adapted to medical segmentation problems by leveraging a large amount of annotated medical data (i.e., computed tomography (CT) scans \cite{kudo2008diagnostic}). However, the generation of reliable large-scale annotations of three-dimensional (3D) medical images requires domain-specific expertise, which is expensive and time-consuming.

\begin{figure}[t]
    \centering
    \includegraphics[width=1\linewidth]{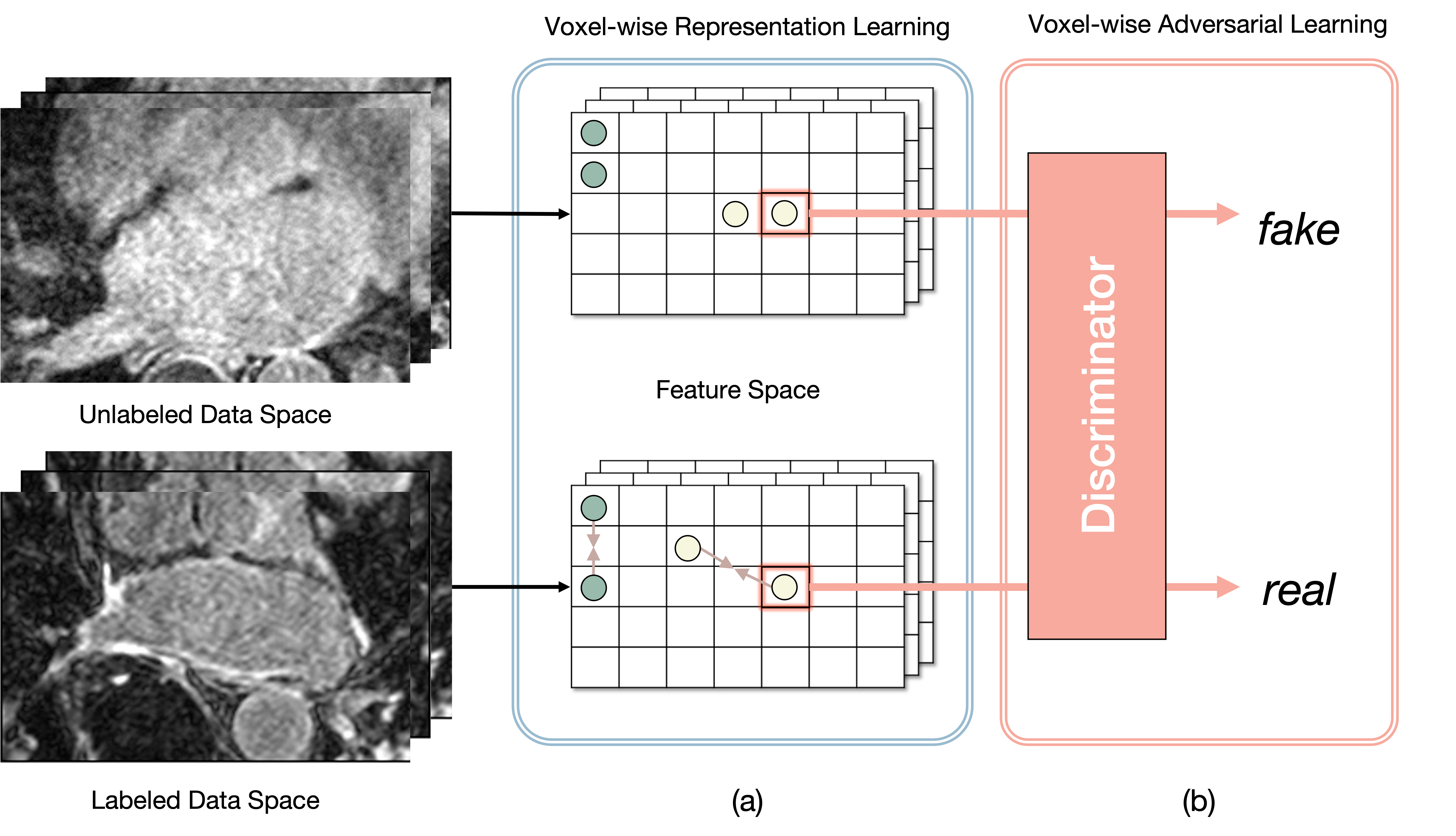}
    \caption{\textbf{Our proposed method} Existing semi-supervised segmentation models learn to map voxels from the data space to the feature space, ignoring global features or class-wise voxel relations. We enforced models to directly learn features representations of labeled and unlabeled data using our proposed method; (a) by defining voxel-wise feature relations of labeled data in the feature space (i.e., voxel-wise representation learning) and (b) by discriminating between the voxel-level features from labeled and unlabeled data (i.e., voxel-wise adversarial learning)}
    \label{fig:summary}
\end{figure}

Significant efforts, such as pretraining, self-supervised learning, and active learning, have been dedicated towards learning from a large number of unlabeled datasets. Semi-supervised learning is one of the approaches used to reduce the annotation cost, where the method simultaneously utilizes a large number of unlableled datasets with a limited number of labeled datasets. The semi-supervised approach assumes that labeled and unlabeled data from the same label share the same or similar underlying distribution (i.e., manifold assumption) \cite{4787647, vanEngelen2019ASO}. We can infer that labeled and unlabeled data usually share similar distributions (e.g., intensities or structures) in medical imaging; consequently, rich semantic information can be embedded using unlabeled data via semi-supervised learning. In practice, several studies on semi-supervised medical image segmentation has proposed effective methods for leveraging unlabeled data. Consistency regularization \cite{tarvainen2018mean}, pseudo-labeling \cite{Leepseudo} and adversarial learning \cite{li2019semi} methods are some of the most commonly used learning methods in semi-supervised learning. The teacher-student model architecture \cite{tarvainen2018mean} has been broadly applied, and it was demonstrated to be effective for consistency regularization- and pseudo-labeling-based methods. Furthermore, improved model performance can be expected through the synergy of representation learning methods from self-supervised and supervised learning by encoding representations from labeled data.

Consistency regularization-based methods \cite{tarvainen2018mean} learn network outputs that are invariant to perturbations or augmentations by adding noise to the unlabeled samples. Different types of methods have been presented to enforce consistency between outputs from different passes, such as uncertainty-aware schemes for data-level consistency \cite{ua_mt} or task-level consistency using a task-transform layer \cite{dtc}. Pseudo-labeling-based methods \cite{MC-Net} generate high-confidence training targets as pseudo-labels for training unlabeled samples. Similar to consistency-based methods, the generated pseudo-labels are used to encourage mutual consistency \cite{MC-Net} to enhance the generalized feature performance. However, these methods learn features by minimizing the loss function in the last layer (i.e., decision space), which can be limited to the local region so that the model learns only the local features of data. On the other hand, adversarial learning-based methods \cite{li2019semi,sassnet} model data distribution of unlabeled samples in an unsupervised setting by utilizing a discriminator. To capture the global shape constraint, a shape-aware adversarial learning method \cite{sassnet} has been proposed for unlabeled data. Although this method is effective for learning shape-aware global features, reproducing features through a separate network is ineffective for learning. Furthermore, both consistency- and adversarial learning- based methods only consider single-class cases and can be limited when they are extended to a multiclass dataset.

Our goal was to improve the representation power for medical image segmentation tasks by leveraging a large amount of unlabeled data. Specifically, we intended to present an effective method that could successfully learn both local and global features from labeled and unlabeled datasets. However, there were several limitations associated with increasing representation power in previous studies. First, these studies focused only on local discrepancy minimization. Most consistency-based methods \cite{ua_mt,MC-Net} calculate output discrepancy in the last layer such that only local features are embedded throughout the training scheme. However, both local and global features should be considered to obtain a better representation space. Second, feature relations across different classes of organs are ignored. Previous studies have only discussed the effectiveness of their methods for single-class data by embedding voxel-to-voxel local relations without distinguishing among different classes. The feature relation between different classes is also important for multiclass data.

In this paper, we propose a novel adversarial learning-based method to incorporate unlabeled data to improve the network performance. We propose a context-aware semi-supervised segmentation method for efficiently learning the distributions of labeled and unlabeled datasets. To resolve the aforementioned problems of recent studies, we considered voxel-wise features from multiple hidden layers, which include both the local and global information of the data, as an input to our voxel-wise feature discriminator to embed distributions of unlabeled datasets. As illustrated in Fig. \ref{fig:summary}b, the job of this discriminator is to determine if a voxel-wise feature belongs to labeled data or unlabeled data (real for labeled data and fake for unlabeled data). This voxel-wise feature discriminator assumes the form of a multitask discriminator that can learn distributions from different classes simultaneously, thereby allowing us to embed class-specific context-aware features in the embedding space. Furthermore, we propose an improved voxel-wise representation learning method (Fig. \ref{fig:summary}a) for labeled data. To effectively embed unlabeled data, we are required to implement well-distributed features from labeled data prior to adversarial learning. In our previous study \cite{lee2022voxel}, we presented an explicit representation learning method for a supervised segmentation task by defining voxel-level feature relations. We adjusted this previous method to embed feature representations from labeled data without information loss using a multiresolution context resizing technique. Moreover, we used the Bootstrap Your Own Latent (BYOL) approach \cite{byol}, instead of SimSiam \cite{chen2020exploring}, for learning stability.

To summarize, our contributions are as follows:
\begin{itemize}
  \item We propose a \textbf{voxel-wise adversarial learning} method that learns both the local and global contexts of labeled and unlabeled data (avoiding the local discrepancy problem) by considering voxel-wise features as an input. Furthermore, our voxel-wise feature discriminator embeds feature relations across different classes by learning a class-specific voxel-wise feature distribution.
  \item We improve the previous \textbf{voxel-wise representation learning} method by overcoming information loss and learning stability problems. This enables our adversarial learning method to effectively learn well-distributed voxel-wise feature representations.
  \item Our method achieves superior results on the Atrial Segmentation Challenge dataset and abdominal multiorgan (MO) dataset when compared with existing state-of-the-art semi-supervised segmentation methods (i.e., consistency regularization, pseudo-labeling and adversarial learning based methods).
\end{itemize}

\par

\section{RELATED WORK} %
\subsection{Semi-Supervised Medical Image Segmentation} 
For semi-supervised medical image segmentation, traditional methods, such as prior- \cite{you2011segmentation} and clustering-based models \cite{portela2014semi}, use hand-crafted features to enhance model performance. With the advanced ability of CNNs, deep learning-based approaches have been widely used for medical image segmentation. Recently, semi-supervised methods based on consistency regularization \cite{ua_mt, dtc}, pseudo labeling\cite{MC-Net}, and adversarial learning-based\cite{dan, sassnet, hung2018adversarial} have proven the effectiveness of incorporating a large amount of unlabeled data for medical image segmentation task.

\textbf{Consistency Regularization.} Consistency regularization is based on the assumption that the segmentation prediction of a network is consistent under realistic perturbations. This motivation was first proposed in \cite{bachman2014learning} and further studied in \cite{laine2017temporal, tarvainen2018mean}. The ${\Pi}$-Model\cite{laine2017temporal} encourages consistent training under different augmentation and dropout conditions. Owing to the noisy training target problem, temporal ensembling \cite{laine2017temporal} adopts the exponential moving average (EMA) of previous evaluations to obtain an ensemble prediction. As a more time-effective method, the teacher-student model \cite{tarvainen2018mean} introduces a pair of networks(i.e., teacher and student networks) and enforces consistency in their predictions. Time efficiency and accuracy can be achieved by averaging model weights, instead of label predictions. 

In medical research, the uncertainty-aware mean teacher (UA-MT) model, proposed in \cite{ua_mt}, utilizes an uncertainty-aware teacher-student framework for LA segmentation. The base model framework was extended from the teacher-student architecture \cite{tarvainen2018mean}, and uncertainty map guidance was adopted to filter out unreliable predictions. More recently, a dual-task consistency (DTC) model \cite{dtc} simultaneously used a pixel-wise segmentation map and level set representation as dual tasks. By utilizing the level set representation, the network could learn the geometric prior. However, the aforementioned methods tend to consider only the local context from the last layer, which can limit the representation of rich global contextual features in the embedding space.

\textbf{Pseudo-labeling.} The concept of pseudo-labeling was proposed in \cite{Leepseudo}, and its variants have presented significant results in semi-supervised learning. For instance, NoisyStudent \cite{noisy} employed a pair of networks, one acting as a teacher and the other as a student. They first trained the teacher network and inferred pseudo-labels for unlabeled images using the teacher network. A larger student network model was then trained using a combination of labeled and pseudo-labeled data, and this process was iterated by converting the student to the teacher. Moreover, a mutual consistency network (MC-Net) \cite{MC-Net} proposed a cycled pseudo-label scheme that used one encoder and two marginally different decoders to utilize unlabeled data. Our method also adopts pseudo-labeling based on teacher-student architecture to infer voxel-wise features from unlabeled data 
in a simple yet powerful manner.

\textbf{Adversarial Learning.} Inspired by the concept of generative adversarial networks  (GANs)\cite{goodfellow2014generative}, several methods that use adversarial learning to exploit unlabeled data have attracted attention in semi-supervised medical image segmentation.
For instance, \cite{2018GANlesion, souly2017semi} used GANs to expand the training set to increase data diversity and avoid overfitting.
Another key idea of using GANs in semi-supervised learning is to force the statistical prior-shape distribution and prediction distribution to be close so that they can effectively learn the distribution on the entire dataset (both labeled and unlabeled).
A shape-aware semi-supervised segmentation network (SASSNET)\cite{sassnet} employs GANs to learn the distribution of both labeled and unlabeled data. This method utilizes the signed distance map (SDM) of images as an input to the discriminator, which plays a vital role in embedding the geometric context of unlabeled data. 
Although this method \cite{sassnet} considers global features employing SDM and a discriminator, context relations between different classes cannot be considered.

\subsection{Representation Learning}
Self-supervised learning methods \cite{ tian2020contrastive, he2020momentum,  chen2020simple} based on contrastive loss have proven to be effective in representation learning. In contrastive learning, positive (similar) pairs are pulled close together, whereas negative (dissimilar) pairs are pushed away.
Because more negative samples can prevent collapse \cite{tian2020contrastive}, several approaches, such as large batch sizes \cite{chen2020simple} or memory banks \cite{he2020momentum}, have been proposed.
Meanwhile, non-contrastive based approaches \cite{byol, chen2020exploring} have shown effective results that avoid collapsing without using negative samples. The BYOL\cite{byol} method is based on teacher-student model, and one branch of momentum encoder enables the network to learn representations without negative samples. Similarly, SimSiam \cite{chen2020exploring} uses a Siamese network and stop-gradient operation, instead of momentum encoder, to prevent collapsing. 

These non-contrastive based approaches can be employed in supervised learning to learn rich representations \cite{lee2022voxel}. Inspired by SimSiam \cite{chen2020exploring}, our previous study \cite{lee2022voxel} presented an effective representation learning method for medical segmentation task by defining voxel-level relations in the embedding space. In this study, we improved our previous method by solving the information loss and learning instability problems of Siamese networks.

\begin{figure*}[h!bt]
    \centering
    \includegraphics[width=\linewidth]{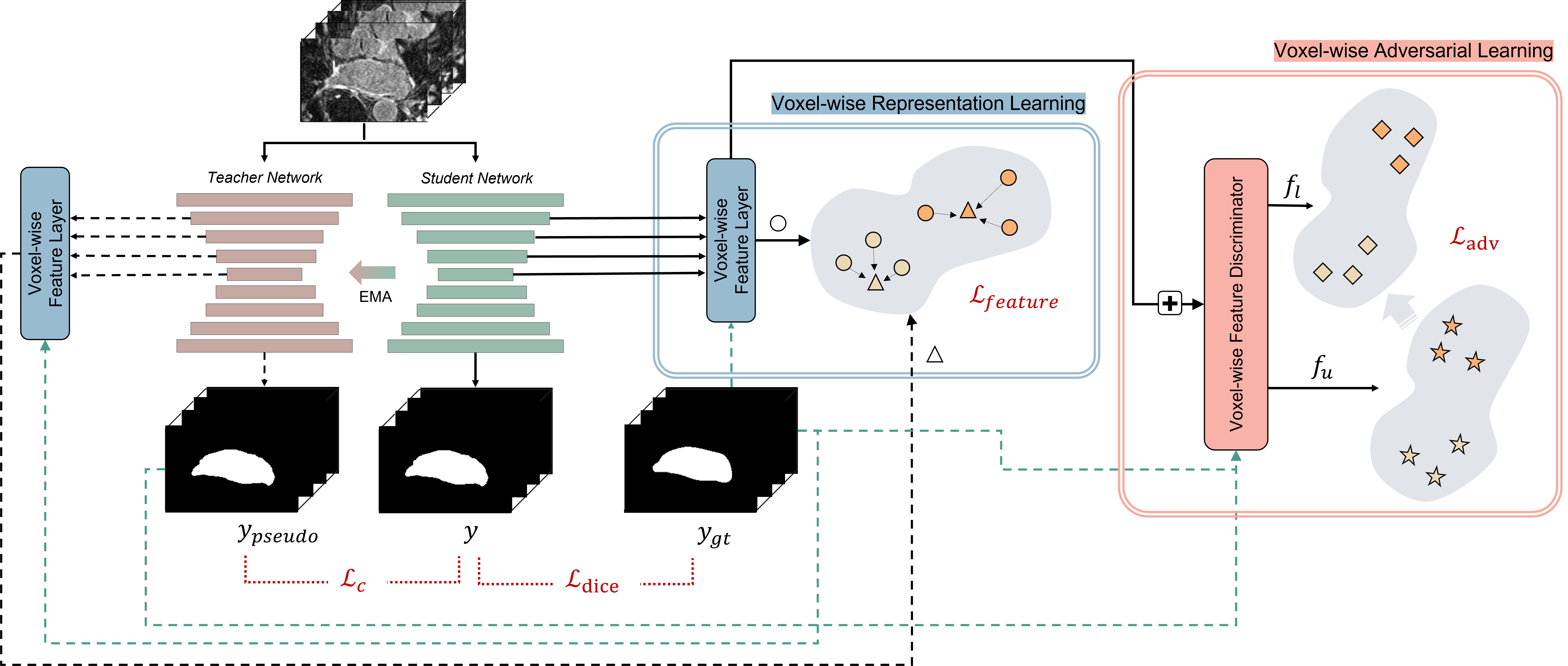}
    \caption{\textbf{Overview of the proposed architecture} Two backbone networks (i.e., VNet \cite{VNet}), i.e., teacher and student networks, take computed tomography scans as an input. The teacher network is learned passively via exponential mean average (EMA). The features ($\bigcirc$) from multiple hidden layers of the student network pass through each section of our proposed network (i.e., voxel-wise feature layer and voxel-wise feature discriminator)so that feature representations from labeled and unlabeled data can be learned. The features ($\triangle$) of the teacher network are used for optimizing the student and our proposed network. The student network is trained using four loss functions ($\mathrm{L}_{dice}$, $\mathrm{L}_{c}$, $\mathrm{L}_{adv}$, and $\mathrm{L}_{feature}$). The gradients are not backpropagated through the dashed lines.}
    \label{fig:architecture}
\end{figure*}

\section{PROPOSED METHOD}

We aim to learn feature representation (i.e., local and global features) from both the labeled and unlabeled datasets. To achieve this, we propose a context-aware semi-supervised segmentation method that can be incorporated into a segmentation network (i.e., VNet \cite{VNet}). The overall architecture of semi-supervised segmentation is illustrated in Fig. \ref{fig:architecture}. There exists a backbone network (i.e., VNet \cite{VNet}) that takes labeled and unlabeled data (i.e., CT scans) as the inputs. We assume a set of training sets containing $N$ labeled data and $M$ unlabeled data, where $N \ll M$. We denote the labeled set as $\mathcal{D}_l = \{(x_i, y_{gt}^i)\}^N_{i=1}$ and unlabeled set as $\mathcal{D}_u = \{(x_i)\}_{i=N+1}^{N+M}$, where $x_i$ represents the 3D volume, and $y_{gt}^i$ denotes the ground-truth label. The proposed architecture for semi-supervised learning consists of two parts: voxel-wise representation learning (the blue box in Fig. \ref{fig:architecture}) and voxel-wise adversarial learning (the red box in Fig. \ref{fig:architecture}). Features from the hidden layers of the backbone network pass through each part to learn feature representations from $\mathcal{D}_l$ and $\mathcal{D}_u$. The voxel-wise adversarial learning method takes voxel-wise features from $\mathcal{D}_l$ and $\mathcal{D}_u$, after which it learns class-specific data distributions. The voxel-wise representation learning method uses voxel-wise features from $\mathcal{D}_l$ and improves current embeddings by defining feature relations from the same class. In Section \ref{sec:adv} and \ref{sec:rl}, we describe the details of these methods. In Section \ref{sec:train}, we explain the overall training process of our proposed method.

\subsection{Voxel-wise Adversarial Learning}
\label{sec:adv}

To leverage a large amount of unlabeled data, the network must be able to learn feature representations using only CT images. Previous consistency-based methods \cite{ua_mt, dtc} have applied a consistency loss function and trained the network for consistent prediction with perturbed or transformed outputs. The consistency loss is computed between $y_{pseudo}$ and $y$ for labeled and unlabeled data. However, this loss is computed in the last layer (i.e., decision space), which embeds only the local features of data. Moreover, it penalizes voxel-wise consistency ignoring class-specific information. It is also problem in \cite{sassnet} that embedded shape-aware global features are only limited to a single class.

\begin{figure}[t]
\begin{center}
     \subfloat[Voxel-wise Feature Discriminator]{\includegraphics[height=2.8in]{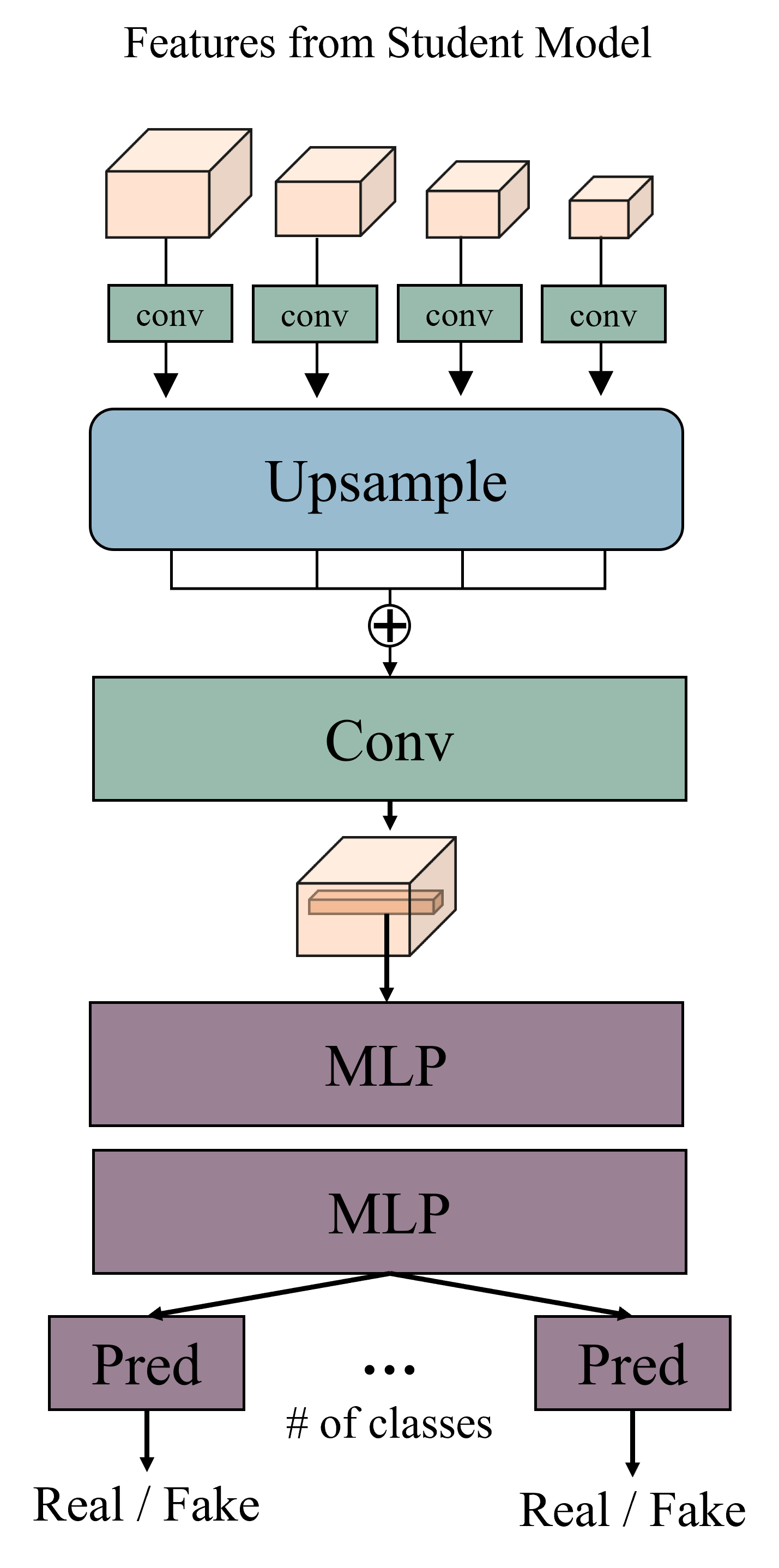} \label{fig:discriminator}}
     \\
    \subfloat[Voxel-wise Feature Layer]{\includegraphics[height=2.8in]{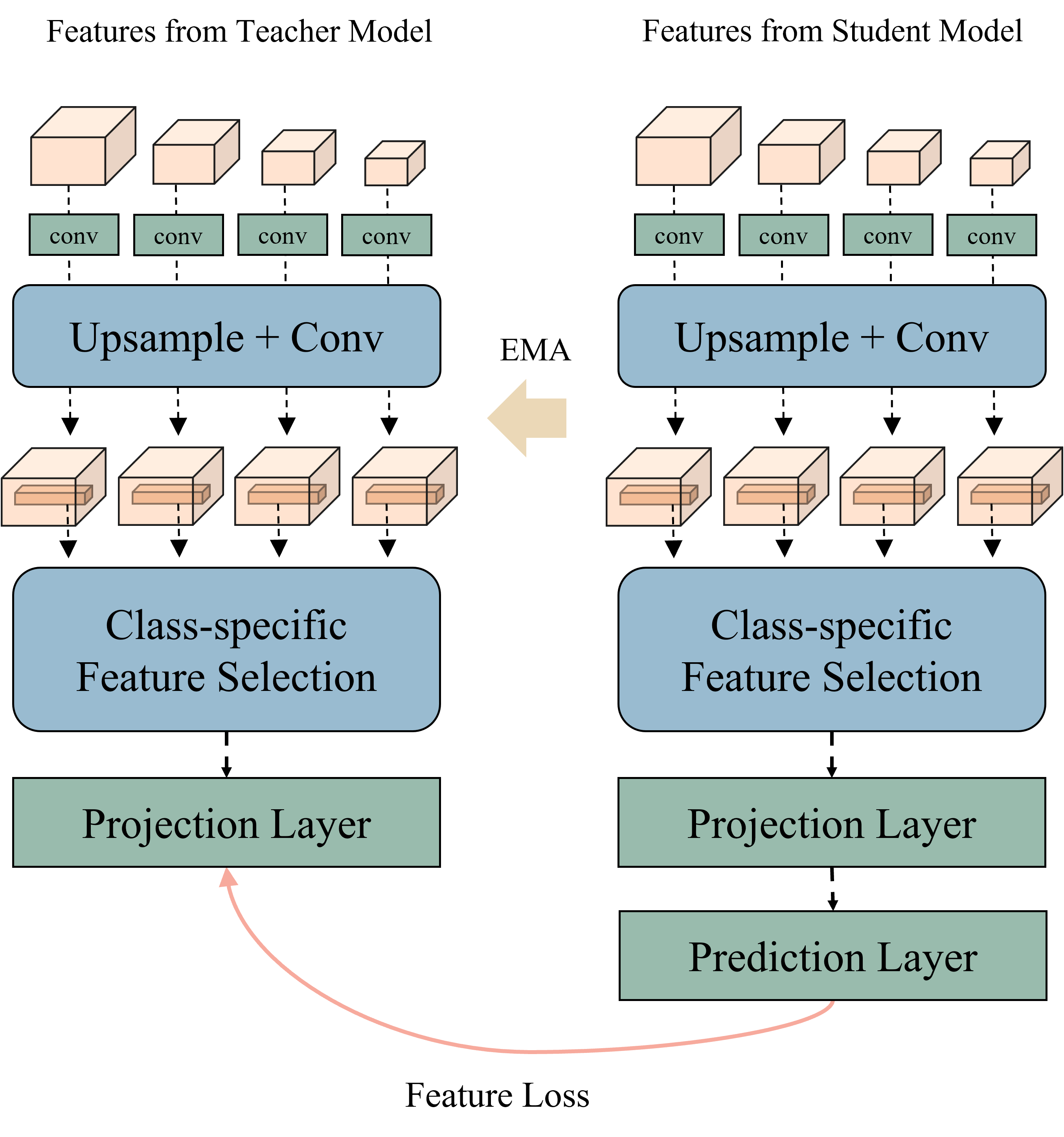} \label{fig:feature}}
    \caption{Details of the proposed architecture. (a) Multiresolution features are fused and the sampled voxel-wise features pass through multilayer perceptron networks (MLPs). The voxel-level features from different classes pass through different prediction layers. This enables the model to learn class-specific voxel-wise distribution of unlabeled data. (b) Multiresolution features pass through multiresolution feature resizing and class-specific feature selection stages. Based on a previous study \cite{byol}, we can learn voxel-wise feature relation in the representation space.}
\end{center}
\end{figure}

To resolve this problem, we propose a novel voxel-wise feature discriminator for embedding class-specific features of both labeled and unlabeled data. As presented in Fig. \ref{fig:discriminator}, our voxel-wise feature discriminator takes a set of multiresolution features, $\{ E(x^1), E(x^2), E(x^3), E(x^4) \}$, as an input, where $E(\cdot)$ denotes an encoder of the backbone, and $E(x^j) \in \mathbb{R}^{H \times W \times D \times C}$ denotes features from the $j^{th}$ hidden layer. These features from multiple hidden layers pass through the convolution layer to adjust the channel size, and each feature is upsampled to the same spatial size. Such features from multiple hidden layers are fused into one by adding an operation and a convolution layer. Thereafter, voxel-level features ($C$-d vector) from this fused feature, $f_{fused}$, pass through a voxel-level feature discriminator, which consists of two multilayer perceptron networks (MLPs) and prediction layer (i.e., linear branch). The number of prediction layers corresponds to the number of class (in case of LA dataset, there exist two classes; foreground and background). The voxel-level features from different classes pass through different prediction layers. To specify the class of each voxel-level feature, we use ground-truth label $y_{gt}$ for labeled data and pseudo-labels $y_{pseudo}$ for unlabeled data, which can be computed using the following equation:
\begin{equation} \label{eq:pseudo}
y_{pseudo} = \mathrm{arg}\mathrm{max} \ \{Teacher(x) > t\},
\end{equation}
where t represents the threshold parameter, which lies in the range of $[0, 1]$.

This different prediction branches enable multiple simultaneous adversarial classification tasks. We define features from labeled data as real and those from unlabeled data as fake so that the encoder of the segmentation network (generator) can generate voxel-level features of unlabeled data with a distribution similar to that of voxel-level features of labeled data. This forces the distributions of class-specific voxel-level features from both labeled and unlabeled features to be close. In this manner, the segmentation network can learn class-specific context-aware features more effectively. The encoder can embed both local and global features using a multiresolution context-fusion technique. In $D$ representing the voxel-wise feature discriminator, we can define our proposed adversarial loss function as follows:
\begin{equation} \label{eq:loss_adv}
\begin{split}
    \mathrm{L}_{adv} = {1 \over N} \sum^N_{n=1} \sum_{f_i \in E(x_n)} log(D(f_i)) \ + \\ 
    {1 \over M} \sum^{N+M}_{m=N+1}\sum_{f_i \in E(x_m)}log(1-D(f_j)).
\end{split}
\end{equation}

\subsection{Voxel-wise Representation Learning}
\label{sec:rl}

In Section \ref{sec:adv}, we propose a new voxel-wise feature discriminator for learning the feature representations of unlabeled data via learning based on the feature distribution of labeled data. In this setting, the most important task is the modeling of the distribution of features from labeled data beforehand. Accurate modeling of the labeled data distribution is essential for effective adversarial learning. In other words, the model is unlikely to learn effectively from adversarial learning if the distribution of labeled data is incorrect. In contrast, the model is likely to learn effectively if distribution is recovered from labeled data. Thus, our model can learn rich feature representations from both labeled and unlabeled data.

In our previous work \cite{lee2022voxel}, we proposed a voxel-level Siamese representation learning method for medical image segmentation tasks. By defining voxel-wise feature relations in the representation space, the model learned feature representations that were effective in the segmentation task. We used the stop-gradient technique and Siamese network from SimSiam \cite{chen2020exploring} to learn voxel-wise feature relations. We also proposed multiresolution feature aggregation method for embedding both local and global features. However, our previous study had two limitations: (1) learning stability and (2) information loss.

In this study, we propose an improved voxel-wise representation learning method for embedding features from labeled data. Inspired by previous studies \cite{byol, lillicrap2015continuous}, we used the learning technique from BYOL \cite{byol}, instead of SimSiam \cite{chen2020exploring}, for the first problem(i.e., learning stability). Using EMA from BYOL enabled the model to produce a more stable prediction target \cite{lillicrap2015continuous} than the stop-gradient technique from SimSiam \cite{chen2020exploring}. As presented in Fig. \ref{fig:feature}, there are teacher and student models; however, the teacher model uses the slow moving average of the student parameter, instead of learning for its own parameter (i.e., EMA). We update the weights of the teacher $\theta_t$ as $\theta_t \gets \lambda \theta_t + (1-\lambda) \theta_s$, where $\lambda$ represents the decay parameter, and $\theta_s$ indicates the weights of the student. Furthermore, for the second problem (i.e., information loss), we propose multiresolution context resizing method. The information loss occurs during the downsampling of mask data to match the class location for each voxel-wise feature. Thus, instead of downsampling the mask data, we upsampled the multiresolution features from the encoder, $E(\cdot)$. Figure \ref{fig:feature} illustrates the upsampling and convolution stage that can reduce information loss.

As explained in Section \ref{sec:adv}, our voxel-wise feature layer (Fig. \ref{fig:architecture} and Fig. \ref{fig:feature}) uses multiresolution features from the encoder of the backbone as an input. These features pass through the upsampling and convolution stages, and voxel-wise features, $p_i^c$, are selected for each class; here, $p_i^c$ refers to the $i^{th}$ voxel-wise feature from class $c$ (class-specific feature selection). These sampled voxel-wise features pass through the projection and prediction layers. The projection layer from the teacher network outputs $z_t$, and the projection and prediction layers from the student network output $p(z_s)$, where $p(\cdot)$ denotes the prediction layer. Based on a previous research \cite{byol}, we used the mean square error between normalized $z_t$ and $p(z_s)$ as the feature loss function. The feature loss function for updating the student network can be defined as follows:
\begin{equation} \label{eq:loss_rl}
    \mathrm{L}_{feature} = {\lVert \bar{p}(z_s) - \bar{z_t} \rVert}^2_2,
\end{equation}
where $\bar{x}$ refers to \textit{l}$_2$-normalization (i.e., $\bar{x} \triangleq {x \over {\lVert x \rVert}_2}$).

\subsection{Training Details}
\label{sec:train}

Our backbone network is based on VNet \cite{VNet}. We first demonstrate a basic VNet \cite{VNet} segmentation training scheme for a labeled dataset. Two VNets \cite{VNet} are displayed in Fig. \ref{fig:architecture}: the teacher and student networks. These two networks take the 3D volume, $x \in \mathbb{R}^{H \times W \times D}$, as an input, and they output prediction masks, $y_{pseudo}$ and $y$ respectively. Based on \cite{ua_mt,sassnet}, we used the dice loss \cite{dc} to maximize the overlap between the ground truth and prediction $y$ to train the student network. We used the labeled dataset (i.e., $\mathcal{D}_l$) to compute the dice loss, which can be defined as
\begin{equation} \label{eq:dice}
    \mathrm{L}_{dice} = \sum^{N}_{i=1}{{2{y_i \cdot Student(x_i)}} \over { {(y_i)}^2 + {(Student(x_i))}^2}}.
\end{equation}
For updating the teacher network, we used the EMA \cite{lillicrap2015continuous} technique.

Following \cite{tarvainen2018mean}, we also added a consistency loss between the softmax predictions of the teacher and student networks for semi-supervised learning. The consistency loss between the outputs of the teacher and student networks can be summarized as follows:
\begin{equation} \label{eq:consistency_loss}
    \mathrm{L}_{c} = \mathbb{E}_x \left[ {\lVert f(x, \theta_t) - f(x, \theta_s) \rVert}^2 \right],
\end{equation}
where $f(\cdot)$ represents the VNet architecture \cite{VNet}. We can stabilize the label prediction by using the teacher-student framework and penalize the predictions that are inconsistent with the target (i.e., output of the teacher network) by adding consistency loss. In this manner, we can learn the generalized local features of both labeled and unlabeled datasets.

The final loss function for training the student network (i.e., VNet \cite{VNet}) is summarized as follows:
\begin{equation} \label{eq:consistency_loss}
    \mathrm{L}_{total} = \alpha \mathrm{L}_{adv} + \beta \mathrm{L}_{feature} + \gamma \mathrm{L}_{c} + \mathrm{L}_{dice}, 
\end{equation}
where $\alpha$, $\beta$ and $\gamma$ represent the coefficients used to balance the different loss terms.

\par

\section{EXPERIMENTAL RESULTS}

\begin{table}
\centering
\caption{Quantitative comparisons of the performances of semi-supervised segmentation models on the left atrium dataset. All models use VNet as the backbone network.}
\label{la_quantative}
\resizebox{\columnwidth}{!}{\begin{tabular}{c|cc|cccc}
\noalign{\smallskip}\noalign{\smallskip}\hline\hline
\multirow{2}{*}{Method} & \multicolumn{2}{c}{\# Scans used} & \multicolumn{4}{c}{Metrics} \\
\cline{2-7}
      & Labeled  & Unlabeled & Dice(\%) & Jaccard(\%) & 95HD(voxel) & ASSD(voxel) \\
\hline
 V\-Net & 8(10\%) & 72 & 79.99 & 58.12 & 21.11 & 5.48 \\
 V\-Net & 16(20\%) & 64 & 86.03 & 76.06 & 14.26 & 3.51 \\
 V\-Net & 80(All) & 0 & 91.14 & 83.82 & 5.75 & 1.52 \\
\hline
 DAP\cite{dap} & 8(10\%) & 72 & 81.89 & 71.23 & 15.81 & 3.80 \\
 UA-MT\cite{ua_mt} & 8(10\%) & 72 & 84.25 & 73.48 & 13.84 & 3.36 \\
 SASSNet\cite{sassnet} & 8(10\%) & 72 & 87.32 & 77.72 & 9.62 & 2.55 \\
 LG-ER-MT\cite{lg-er-mt} & 8(10\%) & 72 & 85.54 & 75.12 & 13.29 & 3.77 \\
 DUWM\cite{duwm} & 8(10\%) & 72 & 85.91 & 75.75 & 12.67 & 3.31 \\
 DTC\cite{dtc} & 8(10\%) & 72 & 86.57 & 76.55 & 14.47 & 3.74 \\
 CVRL\cite{you2021momentum} & 8(10\%) & 72 & 87.72 & 78.29 & 9.34 & 2.23 \\
 MC-Net\cite{MC-Net} & 8(10\%) & 72 & 87.71 & 78.31 & 9.36 & \textbf{2.18} \\
\textbf{Ours} & 8(10\%) & 72 & \textbf{88.42} & \textbf{79.38} & \textbf{8.74} & 2.52 \\
 \hline
DAP & 16(20\%) & 64 & 87.89 & 78.72 & 9.29 & 2.74 \\
UA-MT\cite{ua_mt} & 16(20\%) & 64 & 88.88 & 80.21 & 7.32 & 2.26 \\
SASSNet\cite{sassnet} & 16(20\%) & 64 & 89.54 & 81.24 & 8.24 & 2.20 \\
LG-ER-MT\cite{lg-er-mt} & 16(20\%) & 64 & 89.62 & 81.31 & 7.16 & 2.06 \\
DUWM\cite{duwm} & 16(20\%) & 64 & 89.65 & 81.35 & 7.04 & 2.03 \\
DTC\cite{dtc} & 16(20\%) & 64 & 89.42 & 80.98 & 7.32 & 2.10 \\
CVRL\cite{you2021momentum} & 16(20\%) & 64 & 89.87 & 81.65 & 6.96 & \textbf{1.72} \\
MC-Net\cite{MC-Net} & 16(20\%) & 64 & 90.34 & 82.48 & 6.00 & 1.77 \\
\textbf{Ours} & 16(20\%) & 64 & \textbf{90.56} & \textbf{82.84} & \textbf{5.95} & 1.79 \\
\hline
\hline
\end{tabular}
}
\end{table}

\begin{figure}[]
\begin{center}
     \includegraphics[width=\linewidth]{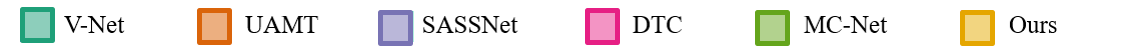}
     \\
    \subfloat[spleen]{\includegraphics[width=\linewidth]{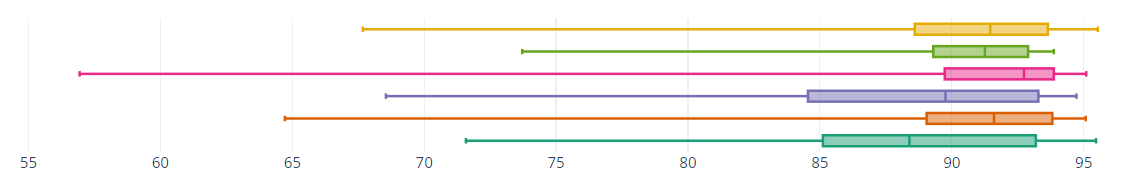}}
    \\
    \subfloat[left kidney]{\includegraphics[width=\linewidth]{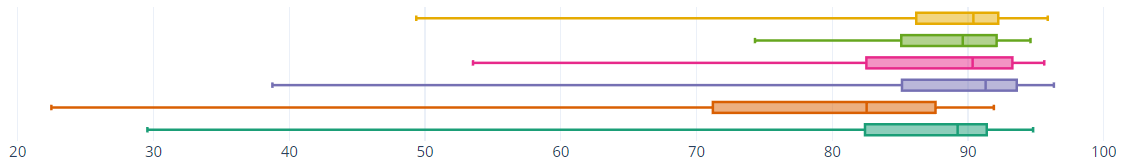}}
    \\
    \subfloat[gallbladder]{\includegraphics[width=\linewidth]{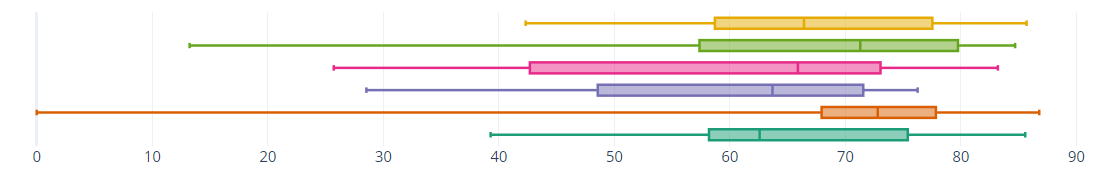}}
    \\
    \subfloat[esophagus]{\includegraphics[width=\linewidth]{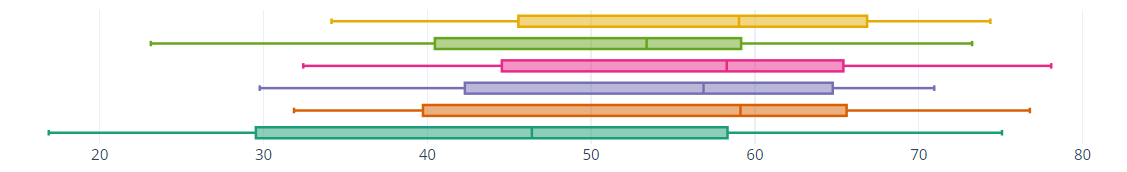}}
    \\
    \subfloat[liver]{\includegraphics[width=\linewidth]{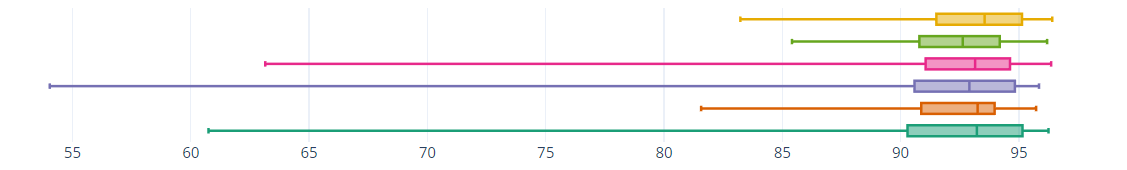}}
    \\
    \subfloat[stomach]{\includegraphics[width=\linewidth]{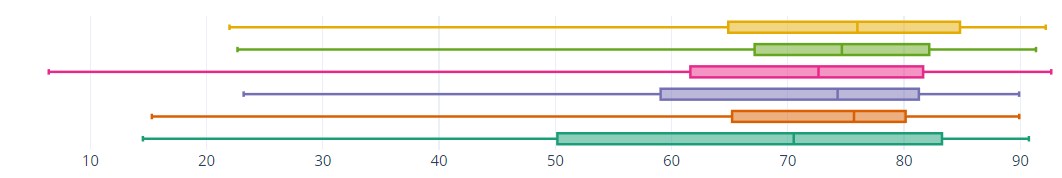}}
    \\
    \subfloat[pancreas]{\includegraphics[width=\linewidth]{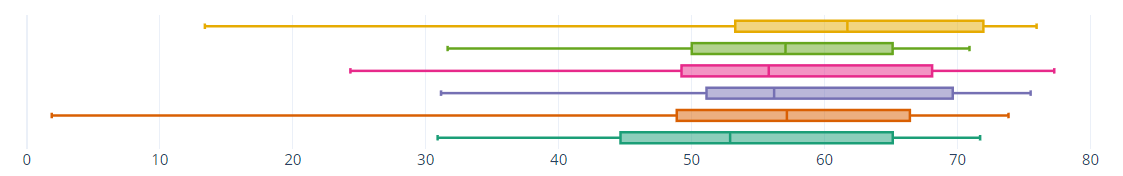}}
    \\
    \subfloat[duodenum]{\includegraphics[width=\linewidth]{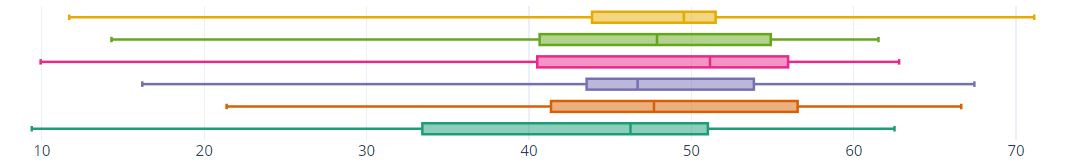}}
    \\
    \caption{Box plots of the dice score coefficient of different methods for eight different organs.}
\label{fig:box_plot}
\end{center}
\end{figure}

\begin{table*}[h!bt]
\centering
\caption{Quantitative comparisons of the performances of semi-supervised segmentation models on the multiorgan dataset}
\label{mo_quantative}
\resizebox{\linewidth}{!}{\begin{tabular}{c|cccc|cccccccc}
\noalign{\smallskip}\noalign{\smallskip}\hline\hline
\multirow{2}{*}{Method} & \multicolumn{4}{c|}{Metrics (average)} & \multicolumn{8}{c}{DSC} \\
\cline{2-13}
      & DSC(\%) & JC(\%) & HD(voxel) & ASSD(voxel)   & spleen & left kidney & gallbladder & esophagus & liver & stomach & pancreas & duodenum \\
\hline
 VNet\cite{VNet} & 66.58   & 54.08 & 5.74 & 1.79 &87.79  & 81.98 &  64.69   &44.88  &91.00  &66.51 &53.39 & 42.42 \\
 UA-MT\cite{ua_mt} & 69.57 & 56.90 & 4.99 & 1.36 & 89.64 & 77.53 & \textbf{67.82} & \textbf{56.19} & 92.21 & 70.73  &54.58  & \textbf{47.86} \\
 SASSNet\cite{sassnet} &69.09  &56.42  &4.85  &1.48 &87.42  &87.26  &60.19  & 54.16 & 90.41 & 69.41 & 57.30  & 46.59 \\
 DTC\cite{dtc} & 69.39 &57.00 &5.78 &1.79 &89.05 &87.03 &59.64 &56.11 &91.23 &68.45 &56.63 &46.99  \\
 MC-Net\cite{MC-Net}&69.76  & 57.34 & 5.61 & 1.90 & 89.15 &\textbf{87.82}  & 64.66 & 50.50 &92.28  &71.22  &56.97  & 45.49 \\
 \textbf{Ours} & \textbf{71.28} & \textbf{59.01} &\textbf{4.32} & \textbf{1.24} & \textbf{89.75} & 87.07 &  66.64 & 56.01 & \textbf{93.03} & \textbf{71.58} & \textbf{59.08} & 47.03 
 \\
\hline
\hline
\end{tabular}
}
\end{table*}

\begin{figure*}[h!bt]
    \centering
    \includegraphics[width=\linewidth]{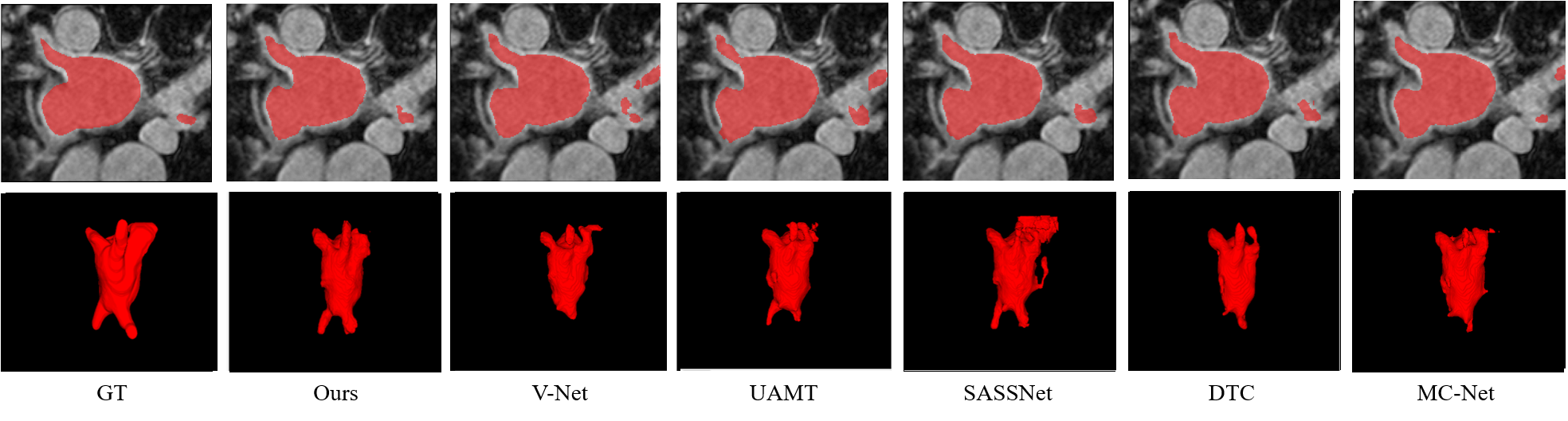}
    \caption{\textbf{Qualitative comparison of different semi-supervised segmentation methods using the left atrium dataset with 20\% labeled data. The first and second rows present the 2D and 3D visualization results, respectively.}}
    \label{fig:Qualitative}
\end{figure*}

\begin{figure*}[h!bt]
    \centering
    \includegraphics[width=\linewidth]{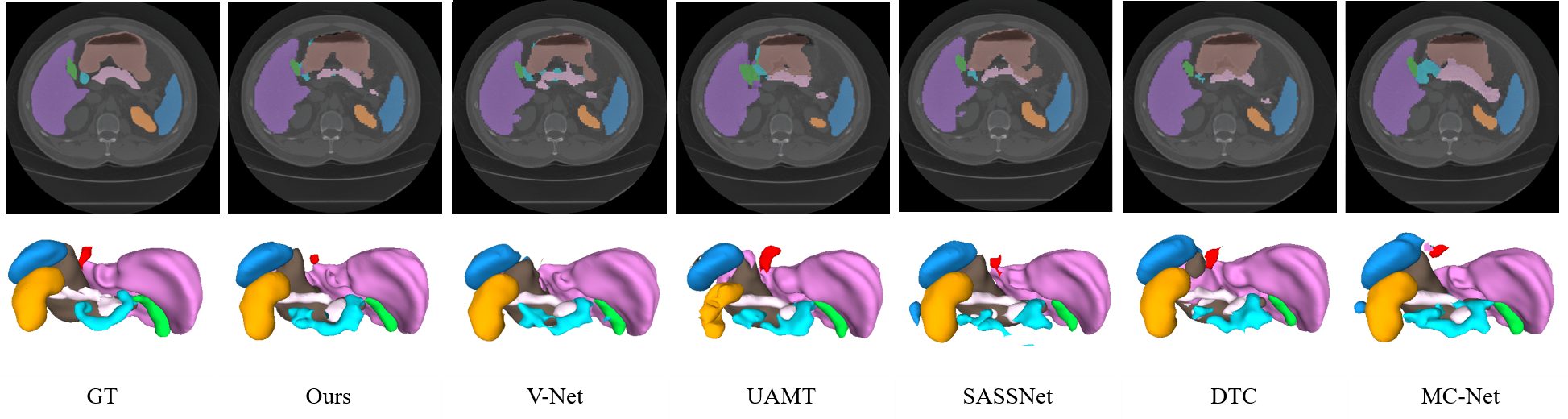}
    \caption{\textbf{Qualitative comparison of different semi-supervised segmentation methods based on the 2D and 3D visualization results obtained using the multiorgan dataset with 20\% labeled data.}}
    \label{fig:mo_all}
\end{figure*}

\subsection{Dataset details}

We evaluated our method using two datasets: the LA dataset from the Atrial Segmentation Challenge and an MO dataset.

\textbf{Atrial Segmentation Challenge dataset}
We used 100 3D gadolinium-enhanced magnetic resonance imaging scans and an LA segmentation mask for training and validation. In the dataset, the scans exhibited an isotropic resolution of $0.6255mm^3 \times 0.6255mm^3 \times 0.625mm^3$. Following the settings of a previous method \cite{ua_mt, sassnet}, the dataset was separated into two sets: training and testing, with 80 images for training and 20 for testing. We applied the same preprocessing method. 

\textbf{Abdominal multiorgan dataset} 
To further evaluate the effectiveness of our method in multiclass segmentation, we evaluated its performance on an MO dataset. We used 90 abdominal CT images: 47 from the Beyond the Cranial Vault dataset \cite{btcv} and 43 from the Pancreas-CT dataset. The segmentation standard consisted of the spleen, left kidney, gallbladder, esophagus, liver, stomach, pancreas, and duodenum. The slice thickness was in the range of $0.5-5.0 mm$ and pixel sizes were in the range of $0.6-1.0 mm$. The dataset was separated into two sets: 70 images for training and 20 for testing. We sampled all abdominal CT images into $128 \times 128 \times 64$ pixels and preprocessed the image using a soft-tissue CT windowing range of $[-200, 250]$ Hounsfield units. After rescaling, we normalized the input images to zero mean and unit variance(i.e., the range of the value is $[0,1]$).

\subsection{Implementation details}
For training both LA and MO dataset, we used a VNet \cite{VNet} architecture as the base network. We set the batch size to 4, and each batch included two labeled patches and two unlabeled patches.

For the LA dataset, we used the stochastic gradient descent optimizer (momentum = 0.9, weight decay of 0.0001) for 6000 iterations, with an initial learning rate of 0.01. The learning rate was divided by 10 for every 2500 iterations. To train the multitask feature discriminator, we followed the method described in \cite{kurach2019largescale}; we used an Adam optimizer ($\beta_1$=0.5, $\beta_2$=0.999) and a learning rate of 0.0002. The weighting parameter $\alpha$ was 0.01 for $\mathrm{L}_{adv}$ and $\beta$ was 0.1 for $\mathrm{L}_{feature}$. Following \cite{sassnet}, we used Gaussian warming-up function $\gamma(t) = 0.001 * e^{-5(1-{t \over t_{max}})^2}$ for consistency loss where $t$ indicates the number of iterations. Based on our previous study\cite{lee2022voxel}, the dimensions of all hidden layers from in voxel-level feature layer were set to 64. Furthermore, we used threshold $t$ of 0.7. We implemented our framework in PyTorch \cite{paszke2019pytorch}, using an NVIDIA TITAN RTX GPU and Tesla V100 GPU. At the inference time, only the VNet framework was used for segmentation.

\begin{table*}[h!bt]
{\tiny
\renewcommand{\arraystretch}{1.5}
\centering
\caption{Visualization of the feature alignment progress during the training phase using our proposed method with ablations and a mutual consistency network. We generated visualization using labeled (marked by triangles) and unlabeled (marked by circles) data, and we present them separately below for comparison.}
\label{tsne_progress}
{\begin{tabular}{c|cc|cc|cc|cc} 
\noalign{\smallskip}\noalign{\smallskip}\hline\hline
\multirow{2}{*}{Iter}  & \multicolumn{2}{c|}{ VNet + $\mathrm{L}_{adv}$}   & \multicolumn{2}{c|}{ VNet +  $\mathrm{L}_{feature}$}   & \multicolumn{2}{c|}{Ours ( VNet +  $\mathrm{L}_{total}$)} & \multicolumn{2}{c}{MC-Net\cite{MC-Net}}                                      \\ 
\cline{2-9} 
&labeled & unlabeled & labeled &  unlabeled  & labeled &  unlabeled  & labeled &  unlabeled   \\ 
\hline
    0.1 k
     &
    \begin{minipage}{.1\textwidth}
    \includegraphics[height=0.5in]{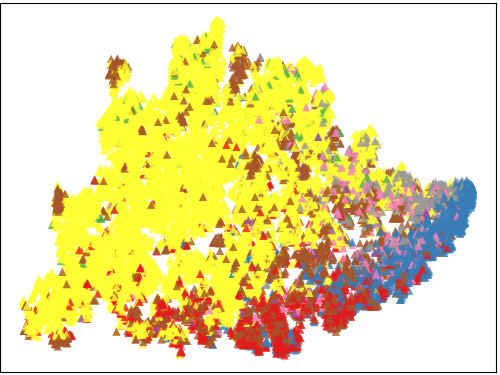}
    \end{minipage}
     &
    \begin{minipage}{.1\textwidth}
    \includegraphics[height=0.5in]{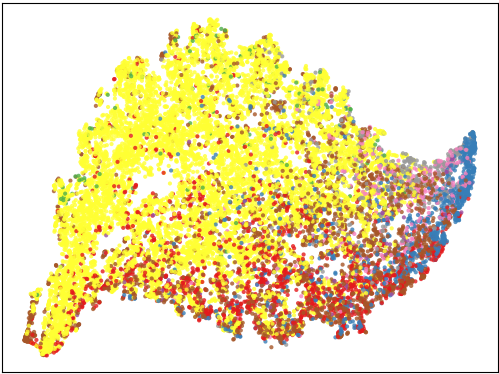}
    \end{minipage}
     &
    \begin{minipage}{.1\textwidth}
    \includegraphics[height=0.5in]{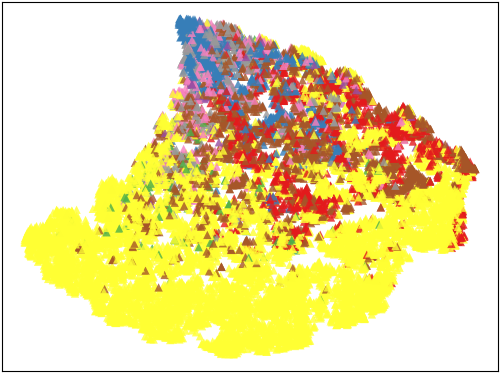}
    \end{minipage}
     &
    \begin{minipage}{.1\textwidth}
    \includegraphics[height=0.5in]{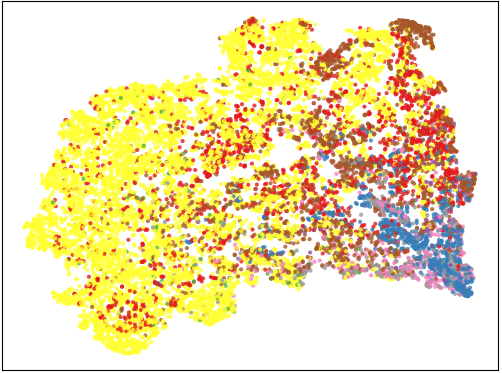}
    \end{minipage}
     &
    \begin{minipage}{.1\textwidth}
    \includegraphics[height=0.5in]{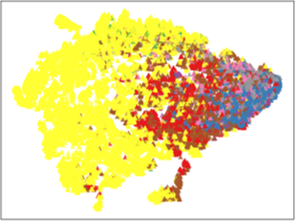}
    \end{minipage}
     &
    \begin{minipage}{.1\textwidth}
    \includegraphics[height=0.5in]{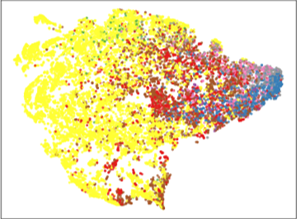}
    \end{minipage}
     &
    \begin{minipage}{.1\textwidth} 
    \includegraphics[height=0.5in]{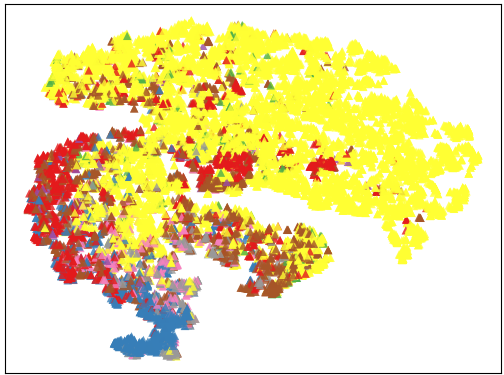}
    \end{minipage}
     &
    \begin{minipage}{.1\textwidth}
    \includegraphics[height=0.5in]{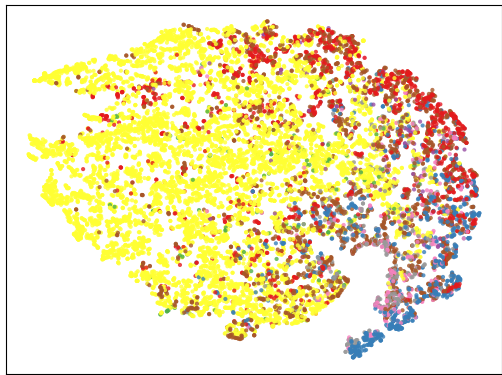}
    \end{minipage}
    
    \\
\hline
    0.5 k
    &
    \begin{minipage}{.1\textwidth}
    \includegraphics[height=0.5in]{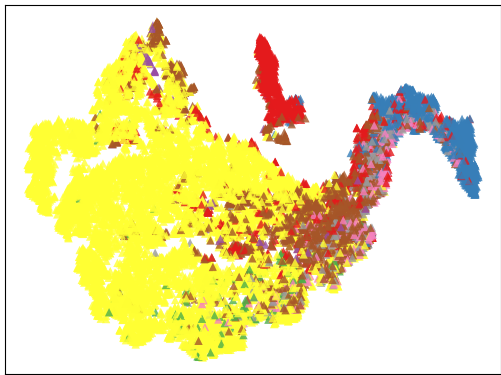}
    \end{minipage}
     &
    \begin{minipage}{.1\textwidth}
    \includegraphics[height=0.5in]{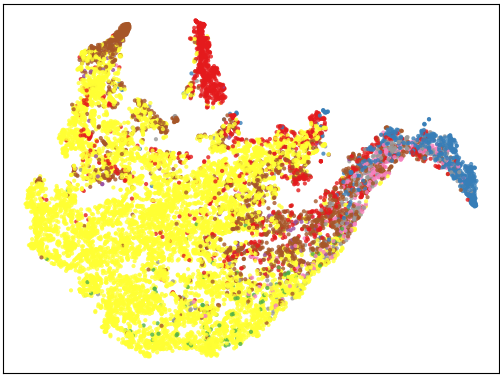}
    \end{minipage}
     &
    \begin{minipage}{.1\textwidth}
    \includegraphics[height=0.5in]{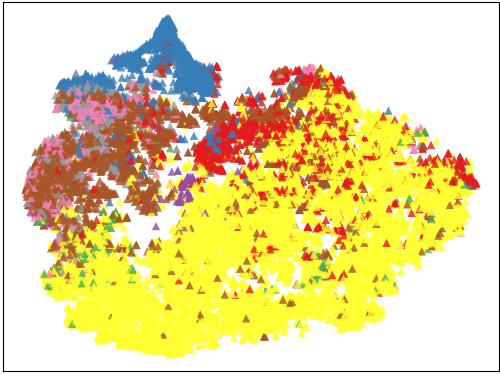}
    \end{minipage}
     &
    \begin{minipage}{.1\textwidth}
    \includegraphics[height=0.5in]{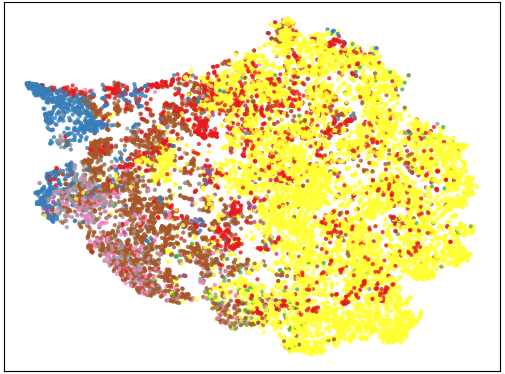}
    \end{minipage}
     &
    \begin{minipage}{.1\textwidth}
    \includegraphics[height=0.5in]{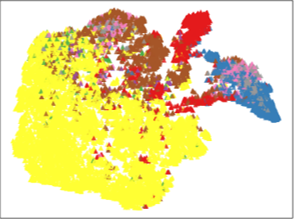}
    \end{minipage}
     &
    \begin{minipage}{.1\textwidth}
    \includegraphics[height=0.5in]{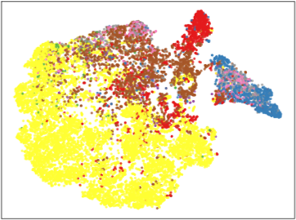}
    \end{minipage}
     &
    \begin{minipage}{.1\textwidth}
    \includegraphics[height=0.5in]{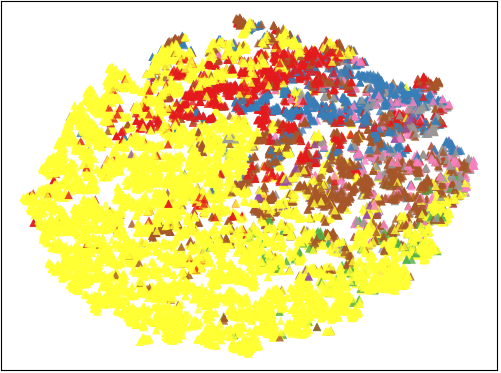}
    \end{minipage}
     &
    \begin{minipage}{.1\textwidth}
    \includegraphics[height=0.5in]{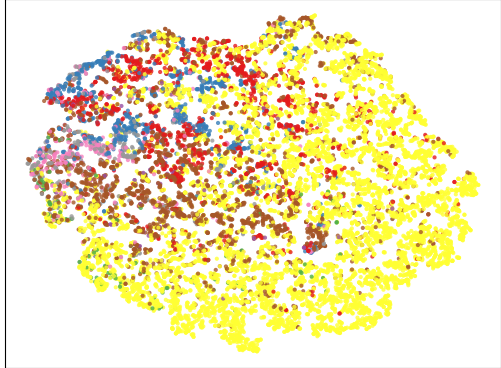}
    \end{minipage}
    \\
\hline

    1 k
     &
    \begin{minipage}{.1\textwidth}
    \includegraphics[height=0.5in]{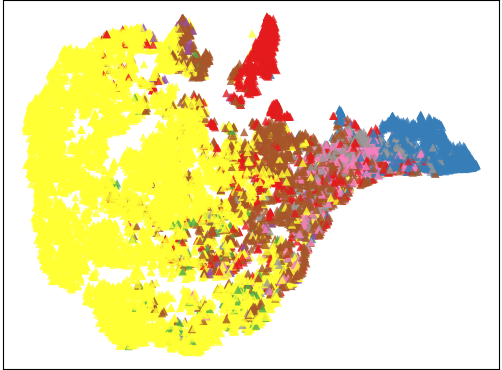}
    \end{minipage}
     &
    \begin{minipage}{.1\textwidth}
    \includegraphics[height=0.5in]{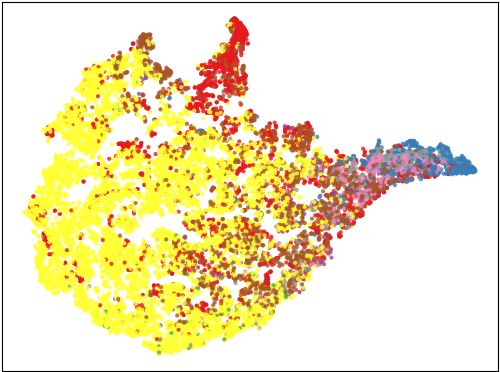}
    \end{minipage}
     &
    \begin{minipage}{.1\textwidth}
    \includegraphics[height=0.5in]{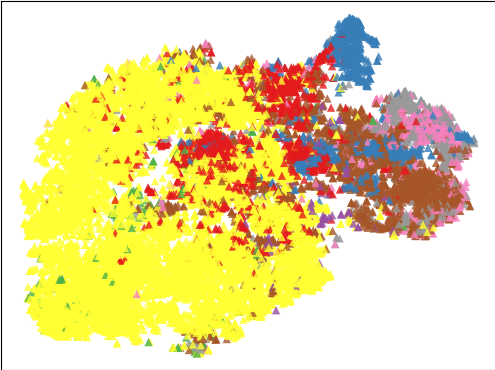}
    \end{minipage}
     &
    \begin{minipage}{.1\textwidth}
    \includegraphics[height=0.5in]{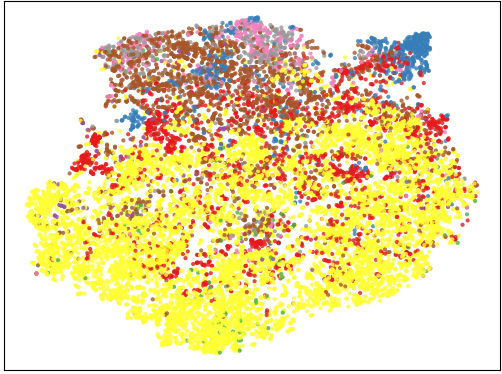}
    \end{minipage}
     &
    \begin{minipage}{.1\textwidth}
    \includegraphics[height=0.5in]{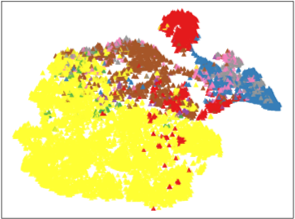}
    \end{minipage}
     &
    \begin{minipage}{.1\textwidth}
    \includegraphics[height=0.5in]{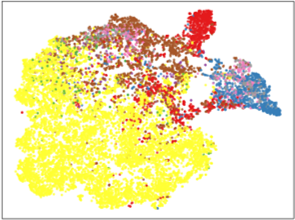}
    \end{minipage}
     &
    \begin{minipage}{.1\textwidth}
    \includegraphics[height=0.5in]{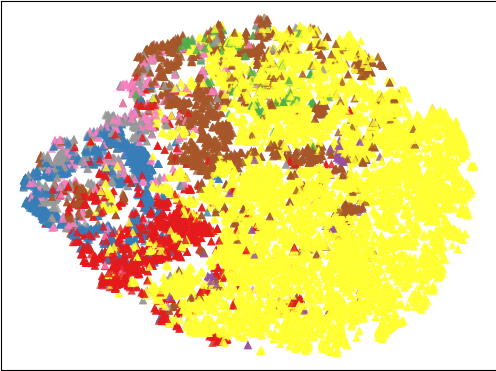}
    \end{minipage}
     &
    \begin{minipage}{.1\textwidth}
    \includegraphics[height=0.5in]{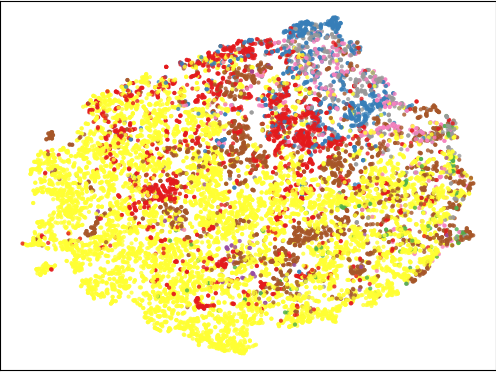}
    \end{minipage}
    \\
    
\hline
\hline
\end{tabular}}
}
\end{table*}

\begin{table}
\centering
\caption{Ablation study of the effectiveness of our proposed method on the left atrium dataset}
\label{table_ablation}
\resizebox{\columnwidth}{!}{\begin{tabular}{c|cc|cccc} 
\noalign{\smallskip}\noalign{\smallskip}\hline\hline
\multirow{2}{*}{Method}                                 & \multicolumn{2}{c|}{ \# Scans used} & \multicolumn{4}{c}{Metrics}                                      \\ 
\cline{2-7}
& Labeled  & Unlabeled               & Dice(\%)       & Jaccard(\%)    & 95HD(voxel)     & ASSD(voxel)        \\ 
\hline
VNet                                                    & 16(20\%) & 64                     & 86.03          & 76.06          & 14.26         & 3.51           \\
VNet+ $\mathrm{L}_{adv}$                                  & 16(20\%) & 64                      & 88.76          & 80.01          & 10.46         & 2.64           \\
VNet+$\mathrm{L}_{feature}$                                 & 16(20\%) & 64                      & 88.67          & 79.85          & 11.52         & 3.31           \\
VNet+$\mathrm{L}_{adv}$+$\mathrm{L}_{feature}$                 & 16(20\%) & 64                      & 90.39          & 82.56          & 10.11         & 2.70           \\
VNet+$\mathrm{L}_{adv}$+$\mathrm{L}_{feature}$+$\mathrm{L}_{c}$ & 16(20\%) & 64                      & \textbf{90.56} & \textbf{82.84} & \textbf{5.95} & \textbf{1.79}  \\
\hline
\hline
\end{tabular}}
\end{table}

For MO dataset, we used Adam optimizer ($\beta_1$=0.9, $\beta_2$=0.999) and an initial learning rate of 0.001 decayed by 0.1 every 2500 iterations. The weighting parameter $\alpha$ was 0.01 for $\mathrm{L}_{adv}$ and $\beta$ was 100 for $\mathrm{L}_{feature}$. The rest of the experimental settings were the same as those employed in the LA dataset experiments.

\subsection{Results}
For our evaluation metrics, we determined the dice score coefficient (DSC) \cite{dc}, Hausdorff distance (HD95; mm) \cite{hd,hd2}, average symmetric surface distance (ASSD; mm) \cite{assd}, and Jaccard Index.

\textbf{Left Atrial Segmentation Challenge dataset.} 
We evaluated the performance of our proposed network in terms of its accuracy by comparing our results with those of the state-of-the-art models, i.e., domain-agnostic prior \cite{dap}, UA-MT\cite{ua_mt}, SASSNet\cite{sassnet}, local and global structure-aware entropy regularized mean teacher \cite{lg-er-mt}, double-uncertainty weighted method \cite{duwm}, DTC\cite{dtc}, contrastive voxel-wise representation learning \cite{you2021momentum}, and MC-Net\cite{MC-Net}. Two semi-supervised settings widely used on the LA dataset were available from a previous study \cite{sassnet} (i.e., using either 10 or 20\% of the labeled data). Table \ref{la_quantative} lists the quantitative results of LA segmentation. The results indicate that our proposed method achieves superior results in terms of the DSC, Jaccard index, and HD95 measurements and achieves competitive results on ASSD under the conditions of both 10\% and 20\% labeled data. 
Qualitative results are illustrated in Fig. \ref{fig:Qualitative}. It can be observed that our method has a higher overlap ratio with respect to the ground truth in both 2D and 3D visualizations, thereby producing fewer false positives.

\textbf{Abdominal multi-organ dataset} To prove the effectiveness of our method on a multiclass dataset, we conducted an experiment on an MO dataset. For comparison, several state-of-the-art models (i.e., UA-MT\cite{ua_mt}, SASSNet\cite{sassnet}, DTC\cite{dtc}, and MC-Net\cite{MC-Net}) and the base network, VNet, were used for evaluation. We considered 20\% of training data among the 70 images as the labeled data (14 labeled) and the others as unlabeled data (54 unlabeled). All the models used VNet as their backbone network. Table \ref{mo_quantative} presents quantitative comparisons of the segmentation results. The results indicate that our method outperforms the other methods in terms of all evaluation metrics (i.e., Dice (71.28\%), Jaccard index (59.01\%), HD (4.32), and ASSD (1.24)). Our method achieves significant improvements in the segmentation of spleen, liver, stomach, and pancreas and demonstrates competitive results for other organs. A box plot for a more precise quantitative comparison is presented in Fig. \ref{fig:box_plot}.
The qualitative results illustrated in Fig. \ref{fig:mo_all} indicate that our method segments multiple organs better than other methods.

\begin{figure}[t]
\begin{center} 
    \subfloat[VNet\cite{VNet}]{\includegraphics[height=1.3in]{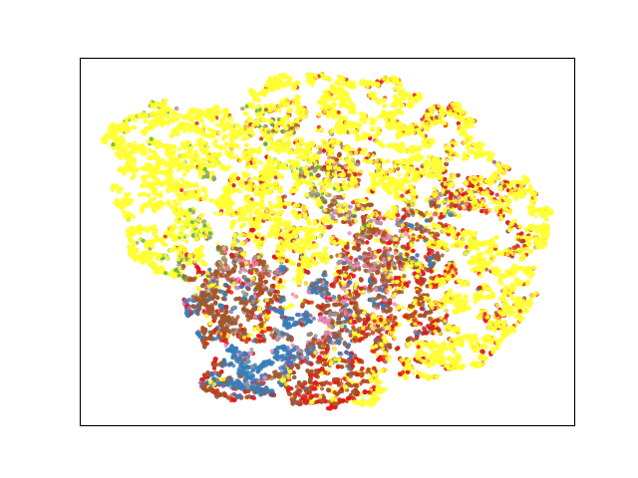} \label{fig:Vnet_tsne}}
    \subfloat[Ours]{\includegraphics[height=1.3in]{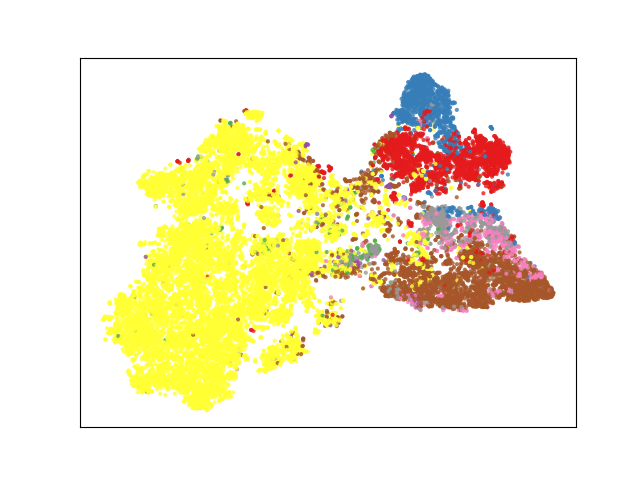} \label{fig:ours_tsne}}
    \caption{Visualization of features from the second layer using (a) VNet\cite{VNet}, (b) our method. The features are colored based on the class labels, and we visualize them using the test dataset (labels are only used for visualization).}
\label{fig:tsne}
\end{center}
\end{figure}


\subsection{Ablation Study} 
We performed an ablation study to investigate the effectiveness of major components of the proposed loss function. We trained VNet under 20\% labeled data using the MO and LA datasets, and the results are listed in Table \ref{tsne_progress} and \ref{table_ablation}, respectively.

From Table \ref{tsne_progress}, we can observe that each major component of our proposed method (i.e., $\mathrm{L}_{adv}$ and $\mathrm{L}_{feature}$) contributes to a more structured representation space in the training process. Specifically, $\mathrm{L}_{adv}$ guides unlabeled data to follow the distribution of labeled data, and $\mathrm{L}_{feature}$ plays a significant role in generating separated feature representation, as expected.

Table \ref{table_ablation} lists the comparison results of the ablations, wherein our losses ($\mathrm{L}_{adv}$, $\mathrm{L}_{feature}$, and  $\mathrm{L}_{c}$) were gradually incorporated. 
The results reveal a significant performance improvement in cases wherein the two losses, $\mathrm{L}_{adv}$ and $\mathrm{L}_{feature}$, were used together, rather than being used separately. This demonstrates that these losses achieve synergy by learning the distribution of unlabeled features from well-distributed labeled features. Furthermore, including the loss function, $\mathrm{L}_{c}$, achieves further improvements by stabilizing label prediction.



\section{DISCUSSION}
Recent semi-supervised segmentation approaches in medical imaging have demonstrated promising results by employing various techniques, such as consistency regularization \cite{ua_mt, dtc}, pseudo-labeling \cite{MC-Net}, and adversarial learning \cite{sassnet}. However, previous methods train the network with the outputs obtained from the final layer, which complicates learning of global features by the network. The proposed method is effective for learning both local and global contexts by embedding voxel-level features with voxel-level feature layers and voxel-level feature discriminators (Table \ref{la_quantative} and Fig. \ref{fig:Qualitative}). We achieved a more structured representation space (Fig. \ref{fig:tsne} b) by defining voxel-level feature (including global and local context) relations in the representation space. On comparison with a previous method \cite{sassnet} which also included global contextual information with the discriminator and SDM, our method achieved superior results (Table \ref{mo_quantative}), particularly for multiclass datasets. By learning class-specific voxel-level features using BYOL\cite{byol} and a multitask discriminator, we achieved a more structured representation space (Fig. \ref{fig:tsne} and Table \ref{tsne_progress}) and precise segmentation results for the multiclass dataset (Table \ref{mo_quantative} and Fig. \ref{fig:mo_all}). This indicates that our method is effective for learning feature relations across different classes. Moreover, as presented in Table \ref{table_ablation}, significant performance improvements can be observed for simultaneous use of the voxel-wise feature discriminator and voxel-wise representation learning; this implies that the unlabeled data distribution follows the labeled data distribution as we intended (Table \ref{tsne_progress}), thereby embedding rich feature representation. In future studies, we can improve the results by suggesting a more efficient method to enable unlabeled data to follow the distribution of labeled data.

\section{CONCLUSION}
In this work, we propose a novel semi-supervised learning method for medical image segmentation tasks. Specifically, our voxel-wise representation learning method embedded feature representations (i.e., local and global features) in the representation space, and our voxel-wise feature discriminator successfully leveraged unlabeled data using the distribution of features from the labeled data. Extensive experimental results indicated that our method achieved competitive results when compared with existing state-of-the-art approaches. Furthermore, our method could provide a more informative representation that embedded class-specific features and achieved superior results in multiclass segmentation. We believe that our approach can provide a useful perspective on medical imaging tasks and can be applied to various medical datasets, regardless of the number of classes.

\section*{Acknowledgements} The authors report no conflict of interest. 


\bibliographystyle{unsrt}
\bibliography{mybib}

\begin{thebibliography}{10}

\bibitem{788580}
Glombitza G, Evers H, Hassfeld S, Engelmann U, and Meinzer HP.
\newblock Virtual surgery in a (tele-)radiology framework.
\newblock {\em IEEE Trans Inf Technol Biomed.}, 3(3):186--196, 1999.

\bibitem{robotics_surgery}
R.~Howe and Y.~Matsuoka.
\newblock Robotics for surgery.
\newblock {\em Annu. Rev. Biomed. Eng}, 1:211–240, 1999.

\bibitem{cad}
van Ginneken~B, Schaefer-Prokop CM, and Prokop M.
\newblock Computer-aided diagnosis: how to move from the laboratory to the
  clinic.
\newblock {\em Radiology}, 261(3):719--732, 2011.

\bibitem{kudo2008diagnostic}
Masatoshi Kudo, Rong~Qin Zheng, Soo~Ryang Kim, Yoshihiro Okabe, Yukio Osaki,
  Hiroko Iijima, Toshinao Itani, Hiroshi Kasugai, Masayuki Kanematsu,
  Katsuyoshi Ito, et~al.
\newblock Diagnostic accuracy of imaging for liver cirrhosis compared to
  histologically proven liver cirrhosis.
\newblock {\em Intervirology}, 51(Suppl. 1):17--26, 2008.

\bibitem{4787647}
O.~Chapelle, B.~Scholkopf, and A.~Zien, Eds.
\newblock Semi-supervised learning (chapelle, o. et al., eds.; 2006)[book
  reviews].
\newblock {\em IEEE Transactions on Neural Networks}, 20(3):542--542, 2009.

\bibitem{vanEngelen2019ASO}
Jesper~E. van Engelen and Holger~H. Hoos.
\newblock A survey on semi-supervised learning.
\newblock {\em Machine Learning}, 109:373--440, 2019.

\bibitem{tarvainen2018mean}
Antti Tarvainen and Harri Valpola.
\newblock Mean teachers are better role models: Weight-averaged consistency
  targets improve semi-supervised deep learning results, 2018.

\bibitem{Leepseudo}
Dong-Hyun Lee.
\newblock Pseudo-label : The simple and efficient semi-supervised learning
  method for deep neural networks.
\newblock {\em ICML 2013 Workshop : Challenges in Representation Learning
  (WREPL)}, 07 2013.

\bibitem{li2019semi}
Wenyuan Li, Zichen Wang, Jiayun Li, Jennifer Polson, William Speier, and
  Corey~W Arnold.
\newblock Semi-supervised learning based on generative adversarial network: a
  comparison between good gan and bad gan approach.
\newblock In {\em CVPR Workshops}, pages 1--11, 2019.

\bibitem{ua_mt}
Lequan Yu, Shujun Wang, Xiaomeng Li, Chi-Wing Fu, and Pheng-Ann Heng.
\newblock Uncertainty-aware self-ensembling model for semi-supervised 3d left
  atrium segmentation, 2019.

\bibitem{dtc}
Xiangde Luo, Jieneng Chen, Tao Song, and Guotai Wang.
\newblock Semi-supervised medical image segmentation through dual-task
  consistency, 2021.

\bibitem{MC-Net}
Yicheng Wu, Minfeng Xu, Zongyuan Ge, Jianfei Cai, and Lei Zhang.
\newblock Semi-supervised left atrium segmentation with mutual consistency
  training.
\newblock {\em CoRR}, abs/2103.02911, 2021.

\bibitem{sassnet}
Shuailin Li, Chuyu Zhang, and Xuming He.
\newblock Shape-aware semi-supervised 3d semantic segmentation for medical
  images.
\newblock {\em Lecture Notes in Computer Science}, page 552–561, 2020.

\bibitem{lee2022voxel}
Chae~Eun Lee, Minyoung Chung, and Yeong-Gil Shin.
\newblock Voxel-level siamese representation learning for abdominal multi-organ
  segmentation.
\newblock {\em Computer Methods and Programs in Biomedicine}, 213:106547, 2022.

\bibitem{byol}
Jean-Bastien Grill, Florian Strub, Florent Altch{\'e}, Corentin Tallec, Pierre
  Richemond, Elena Buchatskaya, Carl Doersch, Bernardo Avila~Pires, Zhaohan
  Guo, Mohammad Gheshlaghi~Azar, et~al.
\newblock Bootstrap your own latent-a new approach to self-supervised learning.
\newblock {\em Advances in Neural Information Processing Systems},
  33:21271--21284, 2020.

\bibitem{chen2020exploring}
Xinlei Chen and Kaiming He.
\newblock Exploring simple siamese representation learning, 2020.

\bibitem{you2011segmentation}
Xinge You, Qinmu Peng, Yuan Yuan, Yiu-ming Cheung, and Jiajia Lei.
\newblock Segmentation of retinal blood vessels using the radial projection and
  semi-supervised approach.
\newblock {\em Pattern recognition}, 44(10-11):2314--2324, 2011.

\bibitem{portela2014semi}
Nara~M Portela, George~DC Cavalcanti, and Tsang~Ing Ren.
\newblock Semi-supervised clustering for mr brain image segmentation.
\newblock {\em Expert Systems with Applications}, 41(4):1492--1497, 2014.

\bibitem{dan}
Yizhe Zhang, Lin Yang, Jianxu Chen, Maridel Fredericksen, David~P Hughes, and
  Danny~Z Chen.
\newblock Deep adversarial networks for biomedical image segmentation utilizing
  unannotated images.
\newblock In {\em International conference on medical image computing and
  computer-assisted intervention}, pages 408--416. Springer, 2017.

\bibitem{hung2018adversarial}
Wei-Chih Hung, Yi-Hsuan Tsai, Yan-Ting Liou, Yen-Yu Lin, and Ming-Hsuan Yang.
\newblock Adversarial learning for semi-supervised semantic segmentation.
\newblock {\em arXiv preprint arXiv:1802.07934}, 2018.

\bibitem{bachman2014learning}
Philip Bachman, Ouais Alsharif, and Doina Precup.
\newblock Learning with pseudo-ensembles.
\newblock {\em Advances in neural information processing systems}, 27, 2014.

\bibitem{laine2017temporal}
Samuli Laine and Timo Aila.
\newblock Temporal ensembling for semi-supervised learning, 2017.

\bibitem{noisy}
Qizhe Xie, Minh-Thang Luong, Eduard Hovy, and Quoc~V. Le.
\newblock Self-training with noisy student improves imagenet classification,
  2020.

\bibitem{goodfellow2014generative}
Ian~J. Goodfellow, Jean Pouget-Abadie, Mehdi Mirza, Bing Xu, David
  Warde-Farley, Sherjil Ozair, Aaron Courville, and Yoshua Bengio.
\newblock Generative adversarial networks, 2014.

\bibitem{2018GANlesion}
Maayan Frid-Adar, Eyal Klang, Michal Amitai, Jacob Goldberger, and Hayit
  Greenspan.
\newblock Synthetic data augmentation using gan for improved liver lesion
  classification.
\newblock In {\em 2018 IEEE 15th International Symposium on Biomedical Imaging
  (ISBI 2018)}, pages 289--293, 2018.

\bibitem{souly2017semi}
Nasim Souly, Concetto Spampinato, and Mubarak Shah.
\newblock Semi supervised semantic segmentation using generative adversarial
  network.
\newblock In {\em 2017 IEEE International Conference on Computer Vision
  (ICCV)}, pages 5689--5697, 2017.

\bibitem{tian2020contrastive}
Yonglong Tian, Dilip Krishnan, and Phillip Isola.
\newblock Contrastive multiview coding.
\newblock In {\em European conference on computer vision}, pages 776--794.
  Springer, 2020.

\bibitem{he2020momentum}
Kaiming He, Haoqi Fan, Yuxin Wu, Saining Xie, and Ross Girshick.
\newblock Momentum contrast for unsupervised visual representation learning.
\newblock In {\em Proceedings of the IEEE/CVF conference on computer vision and
  pattern recognition}, pages 9729--9738, 2020.

\bibitem{chen2020simple}
Ting Chen, Simon Kornblith, Mohammad Norouzi, and Geoffrey Hinton.
\newblock A simple framework for contrastive learning of visual
  representations.
\newblock In {\em International conference on machine learning}, pages
  1597--1607. PMLR, 2020.

\bibitem{VNet}
Fausto Milletari, Nassir Navab, and Seyed{-}Ahmad Ahmadi.
\newblock V-net: Fully convolutional neural networks for volumetric medical
  image segmentation.
\newblock {\em CoRR}, abs/1606.04797, 2016.

\bibitem{lillicrap2015continuous}
Timothy~P Lillicrap, Jonathan~J Hunt, Alexander Pritzel, Nicolas Heess, Tom
  Erez, Yuval Tassa, David Silver, and Daan Wierstra.
\newblock Continuous control with deep reinforcement learning.
\newblock {\em arXiv preprint arXiv:1509.02971}, 2015.

\bibitem{dc}
Carole~H. Sudre, Wenqi Li, Tom Vercauteren, Sebastien Ourselin, and
  M.~Jorge~Cardoso.
\newblock Generalised dice overlap as a deep learning loss function for highly
  unbalanced segmentations.
\newblock {\em Lecture Notes in Computer Science}, page 240–248, 2017.

\bibitem{dap}
Han Zheng, Lanfen Lin, Hongjie Hu, Qiaowei Zhang, Qingqing Chen, Yutaro
  Iwamoto, Xianhua Han, Yen-Wei Chen, Ruofeng Tong, and Jian Wu.
\newblock Semi-supervised segmentation of liver using adversarial learning with
  deep atlas prior.
\newblock In {\em International Conference on Medical Image Computing and
  Computer-Assisted Intervention}, pages 148--156. Springer, 2019.

\bibitem{lg-er-mt}
Wenlong Hang, Wei Feng, Shuang Liang, Lequan Yu, Qiong Wang, Kup-Sze Choi, and
  Jing Qin.
\newblock Local and global structure-aware entropy regularized mean teacher
  model for 3d left atrium segmentation.
\newblock In {\em International Conference on Medical Image Computing and
  Computer-Assisted Intervention}, pages 562--571. Springer, 2020.

\bibitem{duwm}
Yixin Wang, Yao Zhang, Jiang Tian, Cheng Zhong, Zhongchao Shi, Yang Zhang, and
  Zhiqiang He.
\newblock Double-uncertainty weighted method for semi-supervised learning.
\newblock In {\em International Conference on Medical Image Computing and
  Computer-Assisted Intervention}, pages 542--551. Springer, 2020.

\bibitem{you2021momentum}
Chenyu You, Ruihan Zhao, Lawrence Staib, and James~S. Duncan.
\newblock Momentum contrastive voxel-wise representation learning for
  semi-supervised volumetric medical image segmentation, 2021.

\bibitem{btcv}
Eli Gibson, Francesco Giganti, Yipeng Hu, Ester Bonmati, Steve Bandula,
  Kurinchi Gurusamy, Brian Davidson, Stephen~P. Pereira, Matthew~J. Clarkson,
  and Dean~C. Barratt.
\newblock Automatic multi-organ segmentation on abdominal ct with dense
  v-networks.
\newblock {\em IEEE Transactions on Medical Imaging}, 37(8):1822--1834, 2018.

\bibitem{kurach2019largescale}
Karol Kurach, Mario Lucic, Xiaohua Zhai, Marcin Michalski, and Sylvain Gelly.
\newblock A large-scale study on regularization and normalization in gans,
  2019.

\bibitem{paszke2019pytorch}
Adam Paszke, Sam Gross, Francisco Massa, Adam Lerer, James Bradbury, Gregory
  Chanan, Trevor Killeen, Zeming Lin, Natalia Gimelshein, Luca Antiga, Alban
  Desmaison, Andreas Köpf, Edward Yang, Zach DeVito, Martin Raison, Alykhan
  Tejani, Sasank Chilamkurthy, Benoit Steiner, Lu~Fang, Junjie Bai, and Soumith
  Chintala.
\newblock Pytorch: An imperative style, high-performance deep learning library,
  2019.

\bibitem{hd}
D.~P. {Huttenlocher}, G.~A. {Klanderman}, and W.~J. {Rucklidge}.
\newblock Comparing images using the hausdorff distance.
\newblock {\em IEEE Transactions on Pattern Analysis and Machine Intelligence},
  15(9):850--863, 1993.

\bibitem{hd2}
Minyoung Chung, Jingyu Lee, Minkyung Lee, Jeongjin Lee, and Yeong-Gil Shin.
\newblock Deeply self-supervised contour embedded neural network applied to
  liver segmentation.
\newblock {\em Computer Methods and Programs in Biomedicine}, 192:105447, Aug
  2020.

\bibitem{assd}
Fang Lu, Fa~Wu, Peijun Hu, Zhiyi Peng, and Dexing Kong.
\newblock Automatic 3d liver location and segmentation via convolutional neural
  networks and graph cut, 2016.

\end{thebibliography}

\end{document}